# On-Device Continual Learning for Unsupervised Visual Anomaly Detection in Dynamic Manufacturing

**Haoyu Ren[1, 2], Kay Köhle[1,2], Kirill Dorofeev[2], Darko Anicic[2]**

[1] Technical University of Munich, Munich, Germany

[2] Siemens AG, Munich, Germany

## Abstract

In modern manufacturing, Visual Anomaly Detection (VAD) is essential for automated inspection and consistent product quality. Yet, increasingly dynamic and flexible production environments introduce key challenges: First, frequent product changes in small-batch and on-demand manufacturing require rapid model updates. Second, legacy edge hardware lacks the resources to train and run large AI models. Finally, both anomalous and normal training data are often scarce, particularly for newly introduced product variations. We investigate on-device continual learning for unsupervised VAD with localization, extending the PatchCore to incorporate online learning for real-world industrial scenarios. The proposed method leverages a lightweight feature extractor and an incremental coreset update mechanism based on k-center selection, enabling rapid, memory-efficient adaptation from limited data while eliminating costly cloud retraining. Evaluations on an industrial use case are conducted using a testbed designed to emulate flexible production with frequent variant changes in a controlled environment. Our method achieves a 12% AUROC improvement over the baseline, an 80% reduction in memory usage, and faster training compared to batch retraining. These results confirm that our method delivers accurate, resource-efficient, and adaptive VAD suitable for dynamic and smart manufacturing.

## 1.1    Introduction and Background

With the advent of Industrial 4.0, manufacturing is undergoing a shift from long-term, high-volume production towards short-term, small-batch, and flexible operations [1]. The increasing demand for customization is encouraging factories to accommodate frequent changeovers and product families with many variants. However, traditional quality assurance frameworks, which are typically optimized for stable, large-scale production, can struggle to maintain performance under such variability. Visual Anomaly Detection (VAD), a core element of automated inspection in industry made possible by advances in computer vision and deep learning, exemplifies this challenge. In dynamic manufacturing scenarios, each new product variant or revision can introduce subtle visual differences (e.g., shapes, textures, or coatings), resulting in a distributional shift in feature space. This can degrade the accuracy of pre-trained static VAD models, a problem known as data drift. It is necessary to have VAD systems capable of quickly and continuously learning new product variations to sustain reliability. While cloud-based retraining is feasible in theory, it is often impractical due to limited network bandwidth, data privacy concerns, and factory IT policies that restrict external connectivity. These constraints highlight the need for modern VAD systems to not only perform inference, but also learn and update directly at the source where the data is generated.

In practice, most VAD systems are deployed on constrained hardware on the shop floor, such as industrial PCs, embedded systems, or camera-integrated processors, which have limited memory, computational power, and energy budgets. These systems are generally not designed for model training which requires significant computational power. Growing interest is being shown in Edge Artificial Intelligence (Edge AI) and Tiny Machine Learning (TinyML) to enable continual learning directly on constrained devices [2]. These two paradigms aim to make machine learning (ML) more accessible to devices with limited capabilities and shift the ML workload from the cloud to the edge. There have been advancements in hardware [3] [4], software [5] [6], and the ecosystem [7] [8]. For industrial anomaly detection, the implications are transformative. Edge AI and TinyML enable VAD systems to execute efficient inference and on-device model adaptation without relying on cloud infrastructure. This reduces latency, enhances system robustness in the event of intermittent connectivity, and mitigates privacy concerns by keeping sensitive production data within the factory. Furthermore, by fitting the model architecture and learning strategy to the capabilities of existing shop floor hardware, factories can avoid modifying the current hardware topology. This minimises the complexity of integration, the cost of hardware upgrades, and the risk of disrupting established infrastructure.

Despite the recent development of Edge AI and TinyML, a substantial gap remains between research and the practical implementation of applying adaptive VAD on the industrial shop floor. Many existing frameworks, such as [9] [10] [11], depend on large pre-trained backbones or offline batch training pipelines. These approaches are computationally intensive and data-hungry, making them poorly suited for the resource-constrained hardware typically found on the shop floor. Furthermore, edge-oriented methods such as [12] and [13], primarily aim to achieve high detection accuracy in static conditions using benchmark datasets such as MVTec

AD [14] and VisA [15]. They do not reflect the variability and continuous change in customized and dynamic production. As a result, such methods require costly retraining whenever new product variants are introduced. Although continual learning has been studied in the supervised domain [16] [17], its application in unsupervised domain, especially under edge hardware constraints, still lacks discussion. While recent work [18] [19] has evaluated several VAD methods within a continual learning setup, their task designs involve large domain shifts, such as transitioning from detecting anomalies in screws to bottles and cables, or from brain MRIs to liver CTs and chest X-rays. Consequently, their primary focus lies in mitigating catastrophic forgetting, where model performance on earlier tasks drops after adapting to new ones [20]. Nevertheless, these approaches remain relatively naïve, as they rely on offline subsampling that assumes full access to the entire dataset of new tasks. Moreover, the extracted features are simply appended to the memory bank, without considering how knowledge can be transferred or shared across related tasks to enhance subsequent learning. This limitation becomes particularly critical in small-batch production settings, where product variants often exhibit subtle differences and only limited training samples are available. In such cases, leveraging shared representations and inter-task relationships could enhance adaptability and detection performance.

Moreover, as noted by [21], much of the existing research overlooks the operational realities of the industrial environment, often assuming the availability of abundant and diverse training data such as that in the MVTec AD dataset. In practice, however, production environments are tightly controlled, with fixed inspection hardware configurations, including camera, inspection position, and lighting condition. Consequently, all normal images (i.e. images without anomalies) of the same product can appear visually identical, regardless of how many are captured. Collecting additional normal data, therefore, offers little benefit, as it contributes minimal new information. This takes the learning scenario into a one-shot setting, where only a single representative normal sample exists for each product variant. Under these constraints, leveraging prior knowledge across tasks becomes essential to support adaptation to new variants. Furthermore, obtaining anomalous samples, especially labelled ones, is even more challenging while the production plant is operational, particularly when new product variants are introduced. These challenges underscore the need for an efficient, resource-aware continual learning paradigm that can run directly on edge devices using limited training data, while maintaining consistent inspection accuracy in dynamic industrial environments.

To tackle these challenges, we extend PatchCore [22] by integrating efficient on-device continual learning. PatchCore achieves state-of-the-art performance in unsupervised VAD with pixel-level localization across many benchmarks [23] [24]. It leverages pre-trained feature extractors to encode images into a high-dimensional memory bank, known as coreset, which captures the statistical distribution of normal samples without requiring defect samples. During inference, test images are compared against the coreset to determine their similarity using nearest-neighbour search, which enables both detection and localization. Our framework builds on this concept by incorporating incremental and resource-efficient adaptation mechanism. Specifically, we employ a compact feature extractor suitable for constrained hardware and redesign the coreset update mechanism using an incremental k-center selection strategy. This allows the system to perform online, memory-efficient updates by exploiting relationships between embeddings from previous and new tasks. As a result, the model can efficiently adapt to new tasks with minimal data and without the need for full retraining.

We evaluate the proposed method on an industrial use case for dynamic manufacturing using a controlled experimental setup with diverse and interchangeable workpieces that emulate real-world production scenarios. The setup maintains consistent product positioning, camera configurations, and lighting conditions to accurately reflect practical inspection environments. We compare our method with a baseline continual learning approach adopted in [18] that performs coreset subsampling and appends embeddings from new tasks to the existing memory bank. To assess the feasibility of our approach on edge hardware, we deploy and benchmark our method on an NVIDIA Jetson Orin Nano board, using CPU-only execution. The device features an Arm Cortex-A78AE processor and 8 GB of memory, representing a realistic low-power industrial edge computing platform. Experimental results demonstrate that our proposed method delivers substantial gains in both performance and resource efficiency. Specifically, it achieves a 12% increase in AUROC compared to the coreset-subsampling baseline, while reducing memory consumption by over 80% and providing faster learning against the batch training with a batch size of 10.

The rest of the paper is structured as follows: Section 1.2 introduces our method. Section 1.3 covers the use case, describes the experimental settings, and analyzes the results. Finally, Section 1.4 concludes the paper and presents future research directions.

## 1.2 Approach

We propose an on-device continual learning framework that extends PatchCore for fast, localized model adaptation. The method is designed to operate efficiently on resource-constrained edge devices deployed in dynamic environments, enabling a pre-trained PatchCore model to continuously acquire inspection capabilities for new tasks (e.g., product variants) using very limited data, while maintaining performance on previously learned tasks.

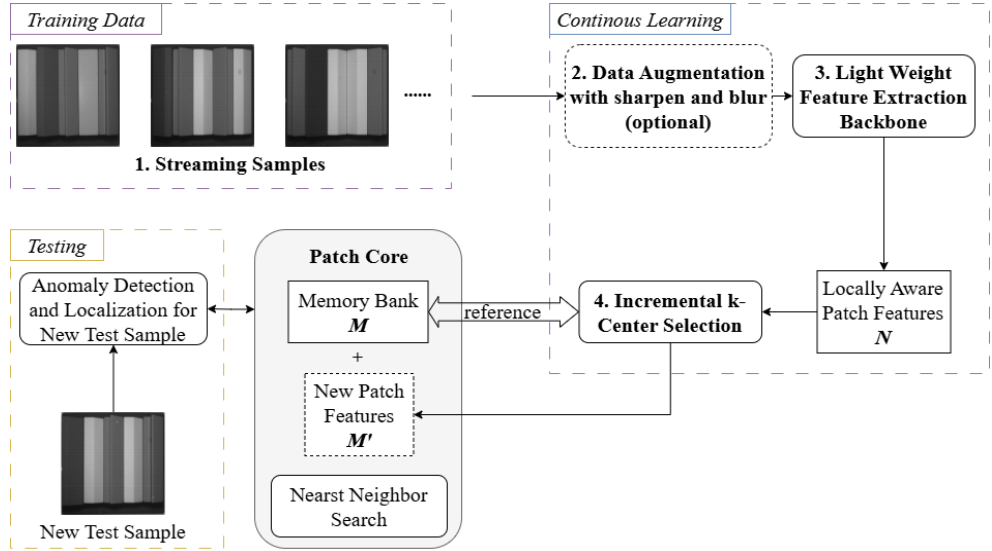

**Figure 1.1** Illustration of our framework extending PatchCore for incremental and resource-efficient adaptation using limited data.

PatchCore is a state-of-the-art, memory-bank-based approach for unsupervised VAD that provides pixel-level localization. It leverages pre-trained feature extractors to transform images into high-dimensional embeddings stored in a memory bank, or coreset. This coreset effectively models the statistical characteristics of normal samples without the need for defective examples during training. At inference, PatchCore computes the similarity between test image features and those in the coreset using a nearest-neighbor search. This allows the system to quantify how "unfamiliar" each local feature is relative to the learned normal distribution, thereby identifying not only whether an image is anomalous but also where anomalies occur with high spatial precision.

While traditional PatchCore is effective, its coreset subsampling is typically performed offline, assuming access to the complete dataset and lacking the ability to handle continual updates without full retraining. As illustrated in Figure 1.1, our framework integrates four key components into PatchCore to enable incremental and resource-efficient adaptation, reducing latency and eliminating the need for a full offline dataset or the transfer of sensitive data beyond the local network. These components are:

- Streaming data consumption in an online learning manner,
- A data augmentation mechanism,
- A lightweight feature extraction backbone, and
- An incremental coreset update mechanism based on k-center selection, incorporating the principles of online learning.

Each component is described in detail below.

### 1.2.1 Streaming Samples

With the concept of online learning, data arrives sequentially, one sample at a time, which aligns well with industrial scenarios where new data, such as samples from new product variants, are continuously captured and provided. Rather than assuming access to large offline datasets, our framework operates in a streaming data setting, where normal samples are processed as they arrive during operation. This paradigm updates the model using one sample per iteration and then discards it, ensuring that only a single data instance is held in memory at any given time. Consequently, this approach significantly reduces memory overhead and enables efficient training even on resource-constrained devices.

### 1.2.2 Data Augmentation

Each incoming image sample undergoes optional data augmentation operations, such as sharpening and blurring, to generate additional samples that enhance the diversity of local texture patterns without altering the overall appearance. Although the original PatchCore method does not employ data augmentation, prior studies [25] have demonstrated that augmentation is crucial for achieving state-of-the-art performance, particularly in scenarios with limited training data. This strategy improves the model's robustness to minor visual variations, such as camera noise, and enhances its generalization to unseen conditions.

### 1.2.3 Lightweight Feature Extractor

After the optional data augmentation step, each image is passed through a lightweight feature extraction backbone. The backbone converts the image into a set of patch-level embeddings that capture its local feature information with normal visual distribution. In this work, a pre-trained lightweight image model, such as MobileNetV3 [26] trained on ImageNet-1K, is employed as the feature extractor. MobileNetV3 contains approximately 3.9 million parameters and is specifically designed to meet the computational and memory constraints of edge hardware, reducing inference latency while maintaining strong representational capacity. Compared to the default Wide-ResNet backbone used in PatchCore, which has approximately 69 million parameters, it offers a more resource-efficient alternative that achieves a balance between efficiency and performance. Depending on the available hardware resources, larger or smaller backbones may also be selected to further optimize feature extraction.

### 1.2.4 Incremental k-Center Selection and Memory Bank Update

Traditional PatchCore constructs a coreset memory bank $M = m^1, m^2, \ldots, m_k$ from normal patch features using offline subsampling, which represents the statistical distribution of normal appearances. While effective, this process assumes access to the entire dataset and cannot accommodate continual learning.

To overcome this problem, we introduce an incremental k-center selection strategy that enables memory-efficient coreset updates incorporating the concept of online learning. Compared to existing continual learning approaches in [18] and [19], where the memory bank is updated in a single step by subsampling standalone patch features from all available data of a new task, our method updates the memory bank continuously by comparing each new data sample with the existing memory bank, one sample at a time. When each new data arrives, a set of locally aware patch features $N$ is extracted using the lightweight backbone. The incremental k-center selection algorithm then identifies the most informative subset of features $M'$ in $N$ by comparing it with the existing memory bank $M$ based on their distances in feature space. Mathematically, each candidate feature $\boldsymbol{p} \in N$ is evaluated using the distance metric

$$d(p, M) = min_{m_i \in M} \|p - m_i\|^2.$$

Features with the highest distance scores are incrementally added to $M'$, which are in the end appended to the memory bank $M$. By referencing prior feature distributions, this continual coreset growth mechanism has two major advantages:

- It maintains representational knowledge from previously seen samples, and

- It efficiently selects diverse new representations with high information content, consuming minimal computational and memory overhead.

During inference, a test sample is processed through the same lightweight feature extractor to generate patch embeddings $t_i \in T$. These embeddings are then compared against the updated coreset using nearest-neighbour search in the feature space

$$s_i = \min_{m_j \in M} |t_i - m_j|_2.$$

The resulting distance $s_i$ represents the anomaly score for patch $i$. Higher scores indicate greater deviation from the learned normal feature distribution. By mapping these patch-level scores back to their spatial positions, a pixel-level anomaly map is obtained, highlighting potential defect regions. The overall image-level anomaly score is computed as either the maximum or average of all patch-level scores, providing a comprehensive measure of the sample's abnormality.

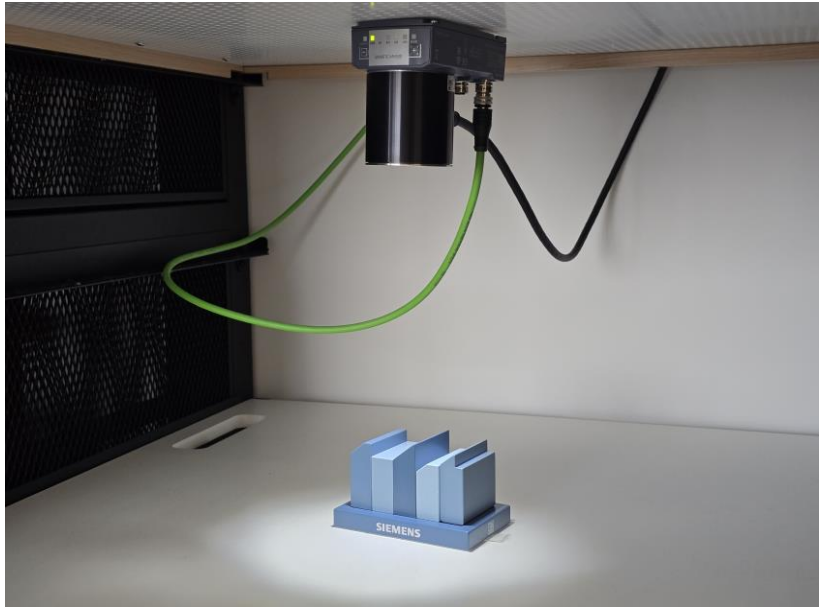

**Figure 1.2** Testbed with a Siemens SIMATIC MV540 camera and five different workpieces on the holder.

## 1.3 Experiments and Evaluation

We quantitatively evaluate our method using an experimental setup that simulates a dynamic manufacturing scenario with frequent product variations. We compare our approach against a continual learning baseline regarding various metrics, such as AUPR, AUROC, memory overhead, and inference latency. To examine the performance at the edge, the method is deployed and benchmarked on an NVIDIA Jetson Orin Nano (CPU-only), representative of low-power industrial edge hardware.

### 1.3.1 Use Case

Our use case focuses on VAD of five workpieces representing manufactured products. The inspection setup, shown in Figure 1.2, consists of a fixed industrial camera (Siemens SIMATIC MV500) mounted above a holder designed to accommodate five workpiece positions. Multiple workpieces with five different shapes are used in the experimental setup. Each workpiece can be freely placed in any of the five positions, with possible repetition and 180° yaw rotation, resulting in more than 50,000 unique combinations. These variations effectively emulate the diversity of product configurations encountered in dynamic, small-batch manufacturing environments.

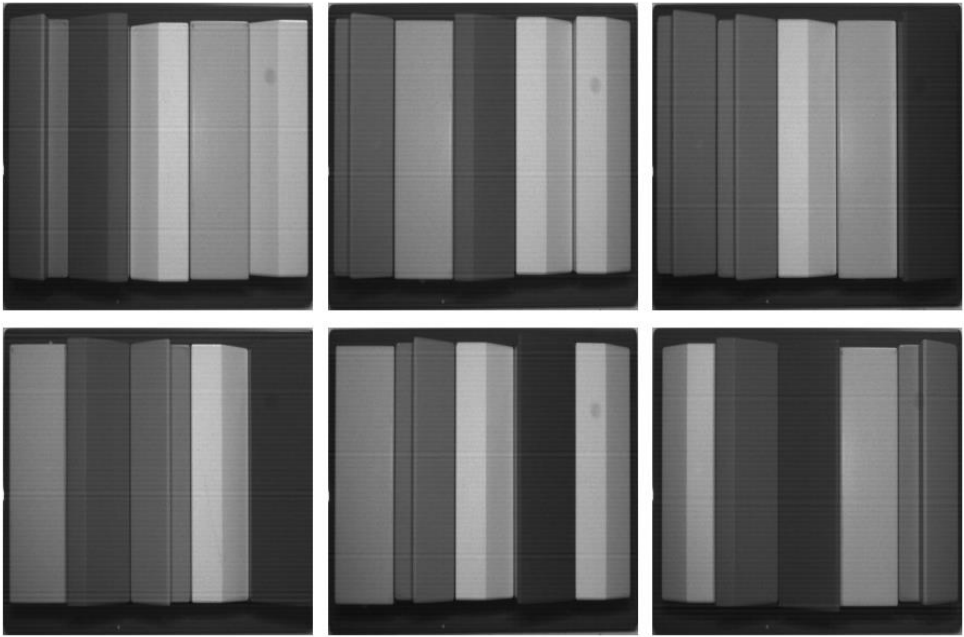

**Figure 1.3** A collection of normal samples of various workpiece combinations.

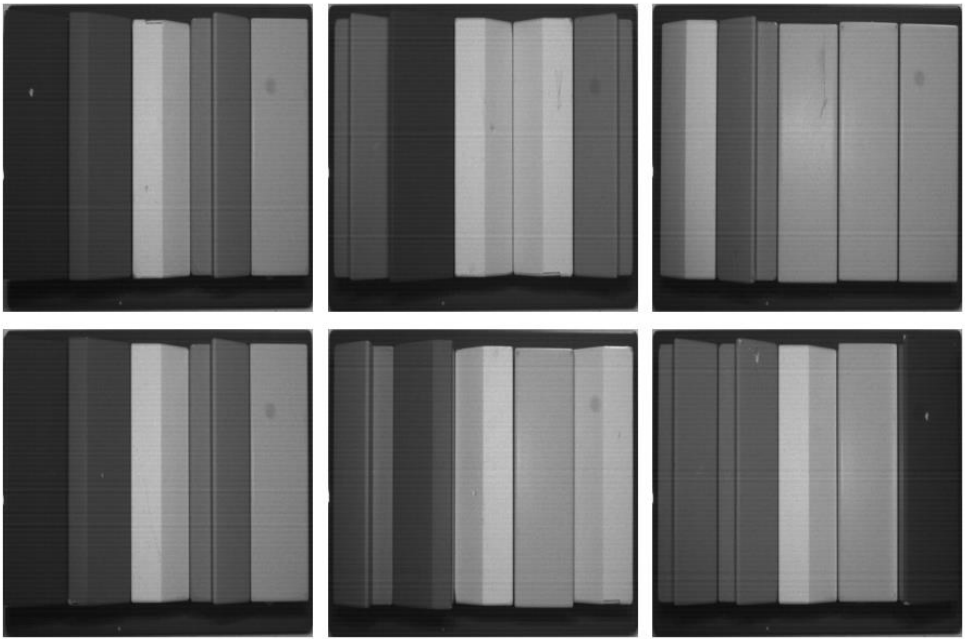

**Figure 1.4** A collection of abnormal samples of various workpiece combinations.

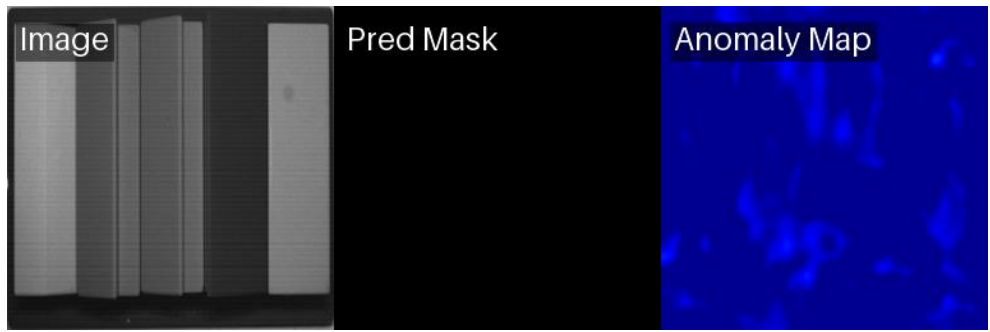

**Figure 1.5** Example of the model prediction on a normal sample.

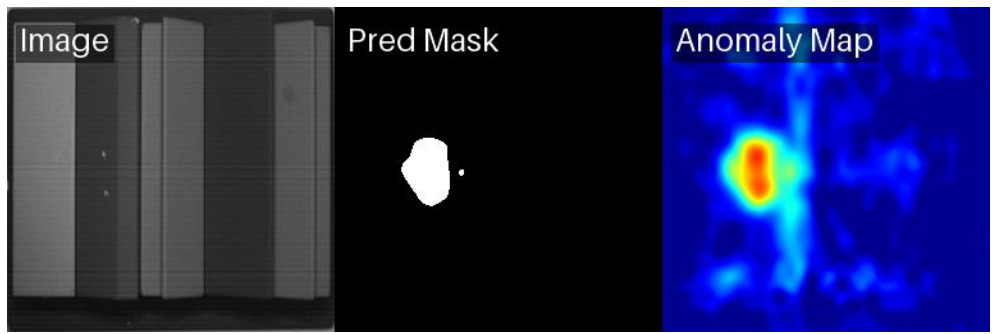

**Figure 1.6** Example of the model prediction on an abnormal sample with highlighted defective region.

Figure 1.3 and Figure 1.4 illustrate representative normal and abnormal workpiece configurations captured by the camera. Given the vast number of potential configurations, training a single VAD model that generalizes across all of them is impractical for edge deployment. Furthermore, collecting extensive image datasets per variant offers limited benefit in this unsupervised setting, as images of the same product variant under such controlled inspection conditions are nearly identical aside from minor camera noise. The key challenge, therefore, lies in enabling rapid on-device continual learning of a pre-trained VAD model to new combinations using only a few, or even a single, normal image per variant.

Figure 1.5 and Figure 1.6 illustrate the prediction results of a trained model on normal and abnormal samples, respectively. These examples demonstrate the capability of the PatchCore model not only to detect abnormal images but also to accurately localize defective regions through anomaly maps, thereby enhancing interpretability and explainability in visual inspection.

### 1.3.2 Experiments Design

We investigate the continual learning capability of a pre-trained PatchCore model when adapting to new workpiece combinations. All experiments start with the same model, trained using the traditional PatchCore method on 117 images, each corresponding to a distinct workpiece configuration. These 117 variants represent standard product lines and already pose a considerable challenge for a single inspection model. Throughout the experiments, the lightweight MobileNetV3 backbone is employed to extract patch-level features.

Subsequently, the model is presented with new workpiece configurations that were not included in the initial training set, where a degradation in detection performance is expected due to data shifts. To address this, we enable continual adaptation by fine-tuning the model online using only a single normal image per new variant. During this process, both the model's performance and the associated hardware requirements are monitored. Here, model performance is evaluated on a fixed dataset comprising 35 unseen workpiece variants, each containing one normal image and multiple abnormal images. The evaluation metrics include the AUPR and the AUROC, which together capture the model's ability to discriminate between normal and abnormal samples under class imbalance, an inherent characteristic of anomaly detection tasks. Hardware efficiency is assessed in terms of memory consumption and inference latency to evaluate the practicality of the approach for edge deployment. We compare our method against a baseline continual learning algorithm from [18], which applies coreset subsampling to extract patch features of given data and append them to the memory bank. Additionally, ablation study is conducted to analyze the impact of key components of our approach, including online adaptation and data augmentation. All experiments are repeated multiple times, and average results are reported to ensure reliability. Hyperparameters are tuned for stable performance rather than for maximal optimization.

### 1.3.3 Evaluation

Figure 1.7 shows the learning progress across different training configurations, showing that the AUPR increases as more normal samples are used for continual learning. A higher AUPR indicates stronger capability in correctly identifying anomalies while reducing false alarms. Here, the baseline shows the slowest improvement, suggesting limited adaptation efficiency. In contrast, our approach incorporating incremental k-center selection achieve faster and more stable performance gains. Also, in the last two configurations, we compare the performance of data augmentation with and without a constraint on the patch feature size. Data augmentation naturally increases the amount of training data, resulting in more features to extract. In the third training configuration, we restrict the number of extracted features added to the memory bank to match that of the second configuration, which does not use augmentation. The results demonstrate that data augmentation can substantially improve model performance, even when the number of extracted features is constrained, effectively balancing accuracy gains with memory overhead.

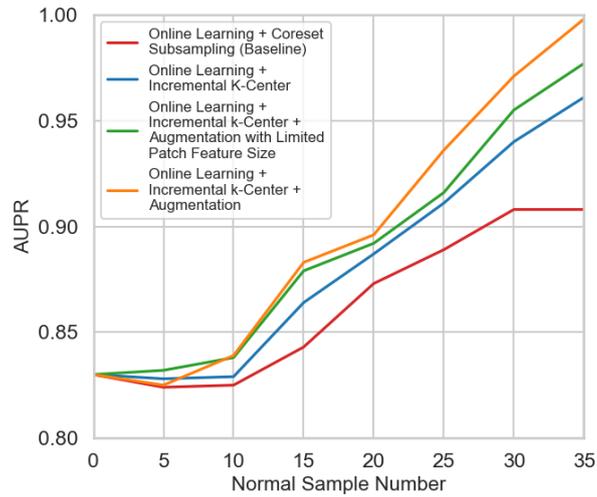

**Figure 1.7** Learning progress of different configurations: AUPR vs. number of normal samples used for finetuning.

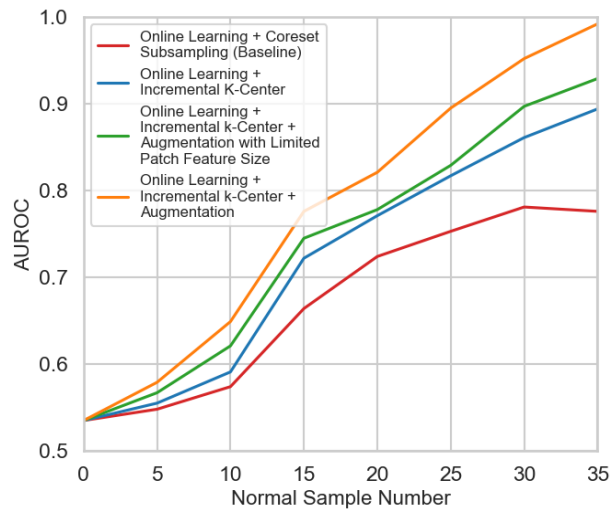

**Figure 1.8** Learning progress of different configurations: AUROC vs. the number of normal samples used for finetuning.

Figure 1.8 shows the AUROC as a function of the number of normal samples used for fine-tuning. AUROC reflects the model's ability to distinguish normal from abnormal samples across various thresholds. A similar trend is observed: methods incorporating incremental k-center sampling and data augmentation achieve the best performance, confirming their effectiveness in improving detection accuracy with limited fine-tuning data.

Table 1.1 summarizes the final detection performance under different training configurations. The "Pre-trained (Lower Bound)" represents the model's initial performance without any fine-tuning, serving as the minimum reference point. The "Pre-trained + BL (Upper Bound)" corresponds to the batch learning setting, where the model is trained with full access to all normal samples, establishing the performance ceiling. Among the online learning configurations, the baseline approach shows moderate improvement over the lower bound. Incorporating incremental k-center selection further narrows the gap. With data augmentation, the performance surpasses that of batch training, even when using a limited patch feature size.

Table 1.2 shows that our online learning methods require significantly less memory than batch learning, reducing usage from around 221 MB to 23 MB for training. The inference stage demands only 0.56 MB, demonstrating that the proposed online learning framework enables efficient training and deployment on memory-constrained devices.

**Table 1.1** Comparison of final detection performance under different configurations, OL: Online Learning, BL: Batch Learning.

| | Pre-trained (Lower Bound) | Pre-trained + BL (Upper Bound) | Pre-trained + OL + Coreset Subsampling (Baseline) | Pre-trained + OL + Incremental K-Center | Pre-trained + OL + Incremental k-Center + Augmentation | Pre-trained + OL + Incremental k-Center + Augmentation with Limited Patch Feature Size |
|---|---|---|---|---|---|---|
| AUPR | 0.829 | 0.972 | 0.908 | 0.961 | **0.998** | 0.977 |
| AUROC | 0.535 | 0.912 | 0.776 | 0.890 | **0.992** | 0.929 |

**Table 1.2** Comparison of memory overhead for training and inference under different configurations, OL: Online Learning, BL: Batch Learning.

| | Training: OL + Coreset Subsampling (baseline) | Training: BL with Batch Size 10 + Incremental K-Center (baseline) | Training: OL + Incremental K-Center | Inference: OL + Incremental K-Center |
|---|---|---|---|---|
| Memory Overhead (MB) | 24.1 | 227.1 | **23.1** | **0.56** |

**Table 1.3** Comparison of memory overhead for storing different numbers of the training samples.

|  | One Image | Three Images | Ten Images |
|---|---|---|---|
| Memory Overhead (MB) | 0.06 | 0.18 | 0.62 |

**Table 1.4** Comparison of average latency for training under different configurations on two platforms.

|  | Training (ours vs. batch learning) | Inference (ours vs. batch learning) |
|---|---|---|
| Laptop – Intel Ultra 5 235U (seconds per epoch) | **0.18** vs. 1.97 | **0.19** vs. 2.02 |
| Jetson Nano - Cortex-A78AE (seconds per epoch) | **0.45** vs. 5.08 | **0.57** vs. 6.34 |

As shown in Table 1.3, our method requires minimal memory for feature storage thanks to the lightweight feature backbone combined with incremental k-center selection. Overhead increases linearly with the number of samples. This demonstrates an efficient balance between memory usage and performance, enabling scalability under constrained resources.

Table 1.4 compares the average training and inference latency across two platforms. Our method achieves substantially lower latency than batch learning, enabling efficient operation even on constrained devices. Moreover, the minimal gap between training and inference times demonstrates the framework's consistency and suitability for real-time applications.

## 1.4    Conclusion and Outlook

We presented a novel on-device continual learning framework for unsupervised VAD in dynamic manufacturing environments. Our approach integrates online learning, a lightweight feature backbone, and incremental k-center memory updates into the standard PatchCore algorithm to achieve strong performance under tight resource budgets. Experimental results show that our method outperforms the coreset subsampling baseline in both accuracy and efficiency. It achieves a 12% improvement in AUROC, reduces memory usage by 80%, and enables significantly faster training. Furthermore, evaluation on a Jetson Nano (CPU-only) confirms its suitability for industrial edge deployment, providing an order-of-magnitude speedup over batch learning and maintaining a minimal training–inference latency gap, with memory overhead below 25 MB for training and below 1 MB for inference.

Looking forward, several avenues can be further explored to strengthen our work. From a practical perspective, future efforts can focus on deploying the framework across additional industrial use cases to validate its generality and robustness.

Methodologically, integrating more advanced continual learning mechanisms, such as enhanced knowledge retention, is a promising way to improve long-term adaptability. In parallel, exploring memory bank optimization techniques such as dimension reduction and quantisation can further reduce the resource requirements of our method. These future advancements aim to broaden the applicability of our approach as an on-device continual learning solution for adaptive VAD in flexible industrial environments.

## Acknowledgements

This work is partially supported by the NEPHELE (ID: 101070487) and SMARTEDGE (ID: 101092908) projects that have received funding from the European Union's Horizon Europe research and innovation program.

## REFERENCES

[1] C. Schmitz, A. Tschiesner, C. Jansen, S. Hallerstede and F. Garms, "Industry 4.0: Capturing value at scale in discrete manufacturing," *McKinsey & Company,* 2019.

[2] V. Tsoukas, A. Gkogkidis, E. Boumpa and A. Kakarountas, "A Review on the emerging technology of TinyML," *ACM Computing Surveys,* vol. 56, p. 1–37, 2024.

[3] A. Khajooei, M. Jamshidi and S. B. Shokouhi, "A super-efficient TinyML processor for the edge metaverse," *Information,* vol. 14, p. 235, 2023.

[4] A. Krishna, S. R. Nudurupati, D. G. Chandana, P. Dwivedi, A. van Schaik, M. Mehendale and C. S. Thakur, "Raman: A reconfigurable and sparse TinyML accelerator for inference on edge," *IEEE Internet of Things Journal,* vol. 11, p. 24831–24845, 2024.

[5] V. J. Reddi, "Generative AI at the Edge: Challenges and Opportunities: The next phase in AI deployment," *Queue,* vol. 23, p. 79–137, 2025.

[6] H. Ren, D. Anicic, X. Li and T. Runkler, "On-device online learning and semantic management of TinyML systems," *ACM Transactions on Embedded Computing Systems,* vol. 23, p. 1–32, 2024.

[7] R. David, J. Duke, A. Jain, V. Janapa Reddi, N. Jeffries, J. Li, N. Kreeger, I. Nappier, M. Natraj, T. Wang and others, "Tensorflow lite micro: Embedded

machine learning for tinyml systems," *Proceedings of machine learning and systems,* vol. 3, p. 800–811, 2021.

[8] R. Kallimani, K. Pai, P. Raghuwanshi, S. Iyer and O. L. A. López, "TinyML: Tools, applications, challenges, and future research directions," *Multimedia Tools and Applications,* vol. 83, p. 29015–29045, 2024.

[9] K. Batzner, L. Heckler and R. König, "Efficientad: Accurate visual anomaly detection at millisecond-level latencies," in *Proceedings of the IEEE/CVF winter conference on applications of computer vision*, 2024.

[10] V. Zavrtanik, M. Kristan and D. Skočaj, "Draem-a discriminatively trained reconstruction embedding for surface anomaly detection," in *Proceedings of the IEEE/CVF international conference on computer vision*, 2021.

[11] X. Du, J. Chen, J. Yu, S. Li and Q. Tan, "Generative adversarial nets for unsupervised outlier detection," *Expert Systems with Applications,* vol. 236, p. 121161, 2024.

[12] M. Barusco, F. Borsatti, D. D. Pezze, F. Paissan, E. Farella and G. A. Susto, "Paste: Improving the efficiency of visual anomaly detection at the edge," in *Proceedings of the Computer Vision and Pattern Recognition Conference*, 2025.

[13] Y. Long, Z. Ling, S. Brook, D. McFarlane and A. Brintrup, "Leveraging unsupervised learning for cost-effective visual anomaly detection," in *IET Conference Proceedings CP885*, 2024.

[14] P. Bergmann, M. Fauser, D. Sattlegger and C. Steger, "MVTec AD–A comprehensive real-world dataset for unsupervised anomaly detection," in *Proceedings of the IEEE/CVF conference on computer vision and pattern recognition*, 2019.

[15] Y. Zou, J. Jeong, L. Pemula, D. Zhang and O. Dabeer, "Spot-the-difference self-supervised pre-training for anomaly detection and segmentation," in *European conference on computer vision*, 2022.

[16] R. Mohandas, M. Southern, E. O'Connell and M. Hayes, "A survey of incremental deep learning for defect detection in manufacturing," *Big Data and Cognitive Computing,* vol. 8, p. 7, 2024.

[17] W. Sun, R. Al Kontar, J. Jin and T.-S. Chang, "A continual learning framework for adaptive defect classification and inspection," *Journal of Quality Technology,* vol. 55, p. 598–614, 2023.

[18] N. Bugarin, J. Bugaric, M. Barusco, D. D. Pezze and G. A. Susto, "Unveiling the anomalies in an ever-changing world: A benchmark for pixel-level

anomaly detection in continual learning," in *Proceedings of the IEEE/CVF Conference on Computer Vision and Pattern Recognition*, 2024.

[19] M. Barusco, F. Borsatti, N. Beda, D. D. Pezze and G. A. Susto, "Towards Continual Visual Anomaly Detection in the Medical Domain," *arXiv preprint arXiv:2508.18013,* 2025.

[20] W. Li, J. Zhan, J. Wang, B. Xia, B.-B. Gao, J. Liu, C. Wang and F. Zheng, "Towards continual adaptation in industrial anomaly detection," in *Proceedings of the 30th ACM International Conference on Multimedia*, 2022.

[21] A. Alzarooni, E. Iqbal, S. U. Khan, S. Javed, B. Moyo and Y. Abdulrahman, "Anomaly detection for industrial applications, its challenges, solutions, and future directions: A review," *arXiv preprint arXiv:2501.11310,* 2025.

[22] K. Roth, L. Pemula, J. Zepeda, B. Schölkopf, T. Brox and P. Gehler, "Towards total recall in industrial anomaly detection," in *Proceedings of the IEEE/CVF conference on computer vision and pattern recognition*, 2022.

[23] Y. Zheng, X. Wang, Y. Qi, W. Li and L. Wu, "Benchmarking unsupervised anomaly detection and localization," *arXiv preprint arXiv:2205.14852,* 2022.

[24] G. Xie, J. Wang, J. Liu, J. Lyu, Y. Liu, C. Wang, F. Zheng and Y. Jin, "IM-IAD: Industrial Image Anomaly Detection Benchmark in Manufacturing," *IEEE Transactions on Cybernetics,* vol. 54, pp. 2720-2733, 2024.

[25] J. Santos, T. Tran and O. Rippel, *Optimizing PatchCore for Few/many-shot Anomaly Detection,* 2023.

[26] A. Howard, M. Sandler, G. Chu, L.-C. Chen, B. Chen, M. Tan, W. Wang, Y. Zhu, R. Pang, V. Vasudevan and others, "Searching for mobilenetv3," in *Proceedings of the IEEE/CVF international conference on computer vision*, 2019.